\documentclass[11pt,a4paper]{article}
\usepackage{times,latexsym}
\usepackage{url}
\usepackage[T1]{fontenc}
\usepackage{graphicx}
\usepackage{amsmath}

\usepackage{multirow}
\usepackage{tacl2021v1}
\usepackage{xspace,mfirstuc,tabulary}

\renewcommand{\anonsubtext}{}
\taclpubformattrue
\iftaclpubformat % this "if" is set by the choice of options

\else

\fi

\title{PRCCF: A Persona-guided Retrieval and Causal-aware Cognitive Filtering Framework for Emotional Support Conversation}

\author{
Yanxin Luo$^{1}$, Xiaoyu Zhang$^{2}$, Jing Li$^{1}$, Yan Gao$^{1}$, Donghong Han$^{1}$\thanks{ \ \ indicates corresponding author.} \\
$^1$School of Computer Science and Engineering, Northeastern University, Shenyang, Liaoning, China \\
$^2$Oracle Cloud Infrastructure, Oracle, Santa Clara, CA, United States \\
Emails: \{2401877, 2110662, gaoyan, handonghong\}@neu.edu.cn, xyz20240703@gmail.com
}

\date{}

\begin{document}
\maketitle
\begin{abstract}
Emotional Support Conversation (ESC) aims to alleviate individual emotional distress by generating empathetic responses. However, existing methods face challenges in effectively supporting deep contextual understanding. To address this issue, we propose PRCCF, a Persona-guided Retrieval and Causality-aware Cognitive Filtering framework. Specifically, the framework incorporates a persona-guided retrieval mechanism that jointly models semantic compatibility and persona alignment to enhance response generation. Furthermore, it employs a causality-aware cognitive filtering module to prioritize causally relevant external knowledge, thereby improving contextual cognitive understanding for emotional reasoning. Extensive experiments on the ESConv dataset demonstrate that PRCCF outperforms state-of-the-art baselines on both automatic metrics and human evaluations. Our code is publicly available at: \href{https://github.com/YancyLyx/ESC}{https://github.com/YancyLyx/PRCCF}.
\end{abstract}

\begin{figure*}[htb]
  \centering
  \includegraphics[width=0.95\textwidth]{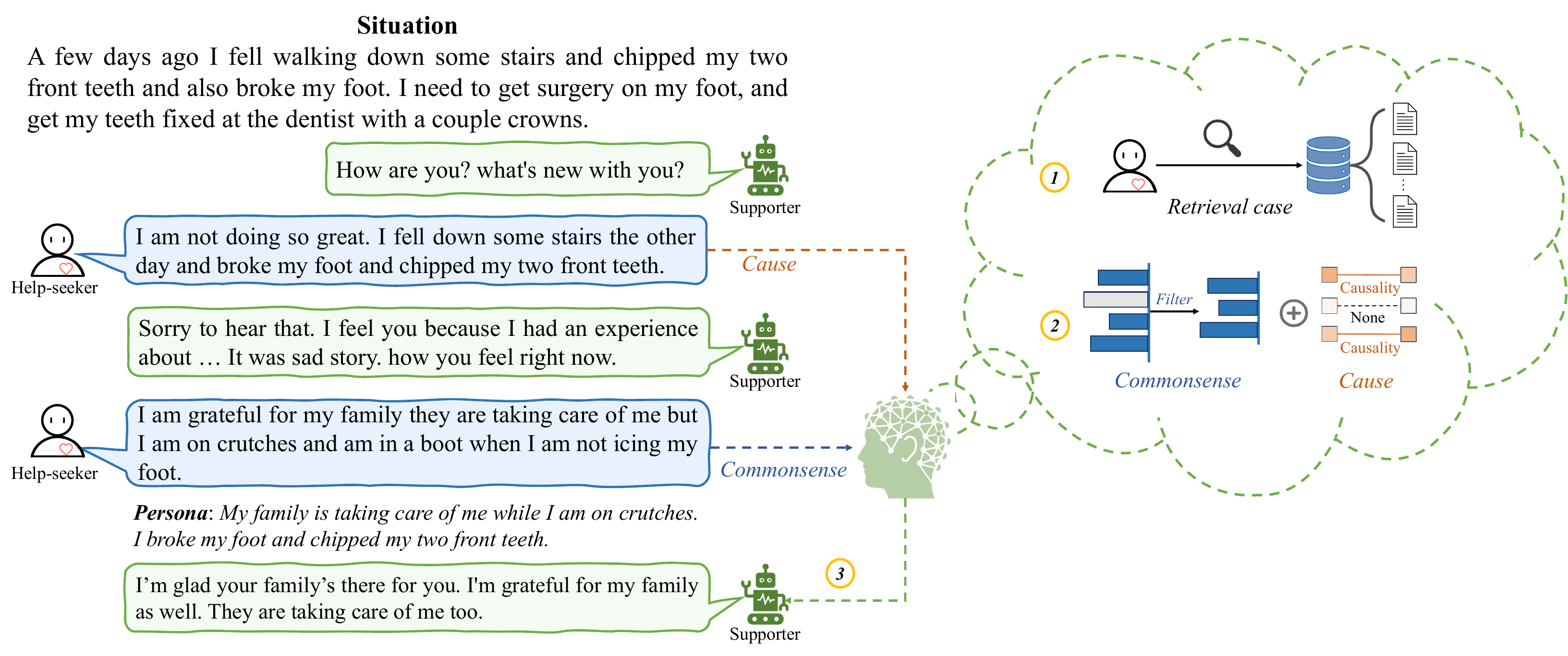} 
  \caption{An example illustrating how PRCCF generates emotional support responses. Persona-aligned demonstrations are retrieved to match the user’s expressive style, while causal cues and filtered commonsense knowledge are integrated as cognitive information to guide empathetic response generation.}
  \label{fig:fig1}
\end{figure*}

\section{Introduction}\label{sec:introduction}

Emotional Support Conversation (ESC) \cite{liu-2021-esconv} aims to alleviate individuals’ emotional distress through empathetic communication, playing an important role in mitigating negative emotions and restoring psychological balance. With the growing prevalence of mental health issues worldwide, the scarcity and high cost of professional psychological services have led to significant gaps in support \cite{olfson2016building,cullen2020mental,vindegaard2020covid}. In this context, emotional support has emerged as an important research area with broad applications in mental health assistance, empathetic customer service, and intelligent companionship \cite{healthsupport,rashkin-2019-empatheticdataset,zhou2018emotional}. However, to fully realize these applications, automated dialogue systems must go beyond recognizing emotional states to understanding their underlying causes and generating genuinely supportive responses, which poses increasing challenges to existing techniques.

In ESC, generating responses aligned with users’ psychological states requires modeling both contextual understanding and personalized characteristics to capture individual emotional needs, which poses challenges for purely generative models in producing consistently empathetic and contextually appropriate responses. According to Hill’s helping skills theory, effective support is developed through learning from concrete cases rather than fixed assumptions \cite{hill2013helping}, motivating the use of retrieval-based demonstrations as reliable guidance for response generation \cite{xu24-ddrcu}. However, due to substantial individual differences, retrieval strategies based solely on semantic similarity often fail to meet users’ actual needs. Although PAL \cite{cheng-etal-2023-pal} incorporates personalized cues, personalization is mainly applied during decoding, leaving demonstration selection coarse and insufficiently aligned with users’ expressive tendencies. Consequently, designing a personalized retrieval mechanism that jointly aligns semantic, emotional, and personal attributes remains a key challenge in ESC.

Beyond personalization, ESC systems also require cognitive understanding of the underlying causes of users’ emotional expressions. Rather than merely recognizing surface-level affective states, systems must infer contextual factors and latent psychological needs to provide genuinely supportive responses. Prior work has incorporated external commonsense knowledge to enhance cognitive empathy \cite{peng2022controlgu-glhg,tu-etal-2022-misc,xu24-ddrcu}, but such approaches typically rely on coarse-grained knowledge integration, limiting their ability to capture causal emotional factors. As a result, generated responses may lack coherence, interpretability, or situational relevance. This underscores the need for fine-grained causal cognition in ESC, enabling systems to identify emotional causes and generate more coherent and supportive responses.

To address the above challenges, we propose PRCCF, a Persona-guided Retrieval and Causality-aware Cognitive Filtering framework for ESC, designed to guide supportive response generation. The framework consists of two core modules: a persona-guided retrieval module and a causality-aware cognitive filtering module. The persona-guided retrieval module selects demonstrations aligned with users’ semantic preferences, emotional tendencies, and expressive styles, providing personalized guidance for response generation. The causality-aware cognitive filtering module refines external knowledge by identifying emotionally relevant causal factors underlying users’ distress, ensuring that the injected knowledge is both relevant and cognitively grounded. 

Figure \ref{fig:fig1} presents an illustrative example of PRCCF in action. In this scenario, the help-seeker experiences emotional distress caused by an unexpected accident involving physical injury and psychological pressure. PRCCF first captures the user’s situation and personalized characteristics to retrieve semantically aligned demonstrations. It then collects causal information and filtered commonsense knowledge to construct cognitively grounded representations. Finally, guided by persona-aligned demonstrations and integrated cognitive information, the model generates a supportive response.

In summary, this paper makes the following contributions:
\begin{enumerate}
\item We propose PRCCF, a novel framework that integrates persona-guided retrieval and causality-aware cognitive filtering to enhance personalization and cognitive grounding in emotional support conversation systems.
\item As core components of PRCCF, we design a persona-guided retrieval module that leverages user characteristics to select demonstrations aligned with users’ semantic preferences and emotional needs.
\item We further introduce a causality-aware cognitive filtering module that extracts emotionally salient causal factors from external knowledge to support coherent and interpretable response generation.
\item Experimental results on the ESConv dataset demonstrate that PRCCF consistently outperforms state-of-the-art baselines in both automatic metrics and human evaluations.
\end{enumerate}

\section{RELATED WORK}

\subsection{Emotional Support Conversation}
ESC aims to provide actionable and supportive guidance to individuals experiencing emotional distress, beyond purely empathetic or surface-level responses \cite{liu-2021-esconv}. To achieve deeper emotional understanding, many studies incorporate external commonsense knowledge (e.g., COMET \cite{bosselut-etal-2019-comet}) to infer users’ emotional challenges and underlying psychological needs \cite{LI2024-DQ-HGAN,peng2022glhg,zhao-etal-2023-transesc,xu24-ddrcu,hao-kong-2025-DKPE}. Some approaches further model emotional transitions across dialogue turns to capture users’ dynamic affective states \cite{zhao-etal-2023-transesc}, while others employ hierarchical structures to connect emotional causality with user intentions through psychological reasoning \cite{peng2022glhg}. In addition, reinforcement learning has been explored to optimize support strategies based on user feedback \cite{tu-etal-2022-misc,zhou-etal-2023-supporter}.

\subsection{Retrieval-enhanced Response Generation}
Retrieval-enhanced response generation has been widely adopted to alleviate issues such as generic responses and hallucinations in dialogue systems \cite{chen-etal-2022-dialogved,li-etal-2016-Dist-n}. Classical methods retrieve demonstrations based on semantic similarity and integrate them via attention mechanisms \cite{shuster-etal-2021-retrieval-augmentation,weston-etal-2018-retrieve}, template-based editing \cite{cai-etal-2019-skeleton}, or controlled rewriting \cite{Wu-2019-response-generation}. To support user-centered generation, recent studies further incorporate user preferences \cite{gupta-etal-2021-controlling} or role consistency constraints \cite{cosplay}.  
However, retrieval-enhanced methods remain underexplored in ESC. Existing approaches such as D${^2}$RCU \cite{xu24-ddrcu} rely primarily on semantic similarity for dynamic demonstration retrieval, which is insufficient to capture multidimensional personalized cues that are crucial for effective emotional support.

\subsection{Commonsense and Causality in Dialogue}
Incorporating external commonsense knowledge has been shown to enhance informativeness, reasoning ability, and contextual coherence in dialogue systems \cite{bosselut-etal-2019-comet}. For ESC, causal reasoning is particularly important, as it enables the system to identify emotional triggers, understand underlying psychological mechanisms, and generate targeted support \cite{chen2024CauESC,hao-kong-2025-DKPE}. For instance, CauESC \cite{chen2024CauESC} employs cause-aware attention to emphasize knowledge, while DKPE \cite{hao-kong-2025-DKPE} predicts emotional causes to filter external knowledge.  
Nevertheless, many existing methods inject knowledge without sufficiently verifying causal relevance, which may introduce noise, degrade reasoning quality, and weaken contextual coherence and support effectiveness.

\begin{figure*}[htbp]
  \centering
  \includegraphics[width=0.95\textwidth]{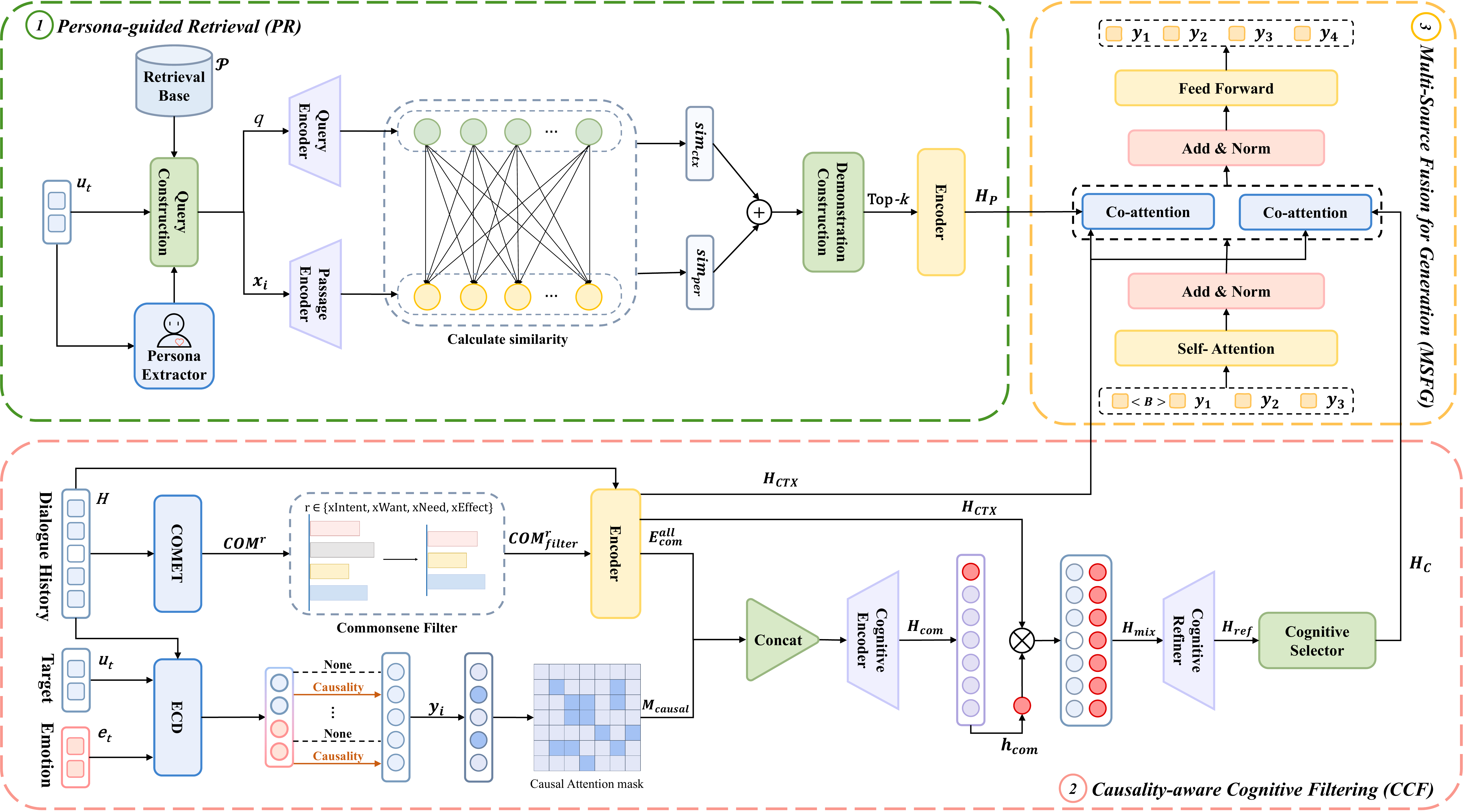} 
  \caption{The figure illustrates PRCCF framework, consisting of three main modules: the Persona-guided Multi-View Retriever, the Causality-aware Cognitive Filtering, and the Multi-Source Fusion for Generation.}
  \label{fig:framework}
\end{figure*}

\section{Methodology}
\subsection{ESC Task Formulation}
Given a multi-turn dialogue context ${H}=\{u_1,u_2,\dots,u_{t-1}\}$ and the help-seeker's current utterance $u_t$, the goal of ESC is to generate a supportive response $y=\{y_1,y_2,\dots,y_n\}$ that is empathetic and contextually appropriate. Following prior work, we also consider the historical support strategy sequence ${S}=\{s_1,s_2,\dots,s_{t-1}\}$ as auxiliary guidance. Formally, the objective is to model the conditional probability $P(y \mid H,u_t,S)$.

\subsection{Overview of PRCCF}
To address the challenges of personalization and cognitive grounding in ESC, we propose PRCCF, which extends the standard conditional generation framework with two core components: a Persona-guided Retrieval (PR) module and a Causality-aware Cognitive Filtering (CCF) module. The overall architecture is illustrated in Figure \ref{fig:framework}.

Finally, the retrieved demonstrations ${P}$ and the causally filtered knowledge ${C}$ are incorporated into the generator, resulting in an enhanced conditional objective:$P(y \mid H, u_t, S, P, C)$,
which enables PRCCF to generate responses that are both emotionally supportive and contextually coherent.

\subsection{Persona-guided Retrieval}

To generate empathetic and personalized responses in ESC, we introduce a Persona-guided Retrieval (PR) module that retrieves demonstration examples by jointly modeling semantic relevance and persona consistency.

\textbf{Retrieval Corpus.}
We construct a retrieval corpus $\mathcal{P}$ from the ESConv \cite{liu-2021-esconv} training set, where each entry is represented as:
\begin{equation}
\mathcal{P} = ( u_i, s_i, r_i, p_i ),
\end{equation}
with $u_i$ denoting the help-seeker utterance, $s_i$ the support strategy, $r_i$ the supporter response, and $p_i$ the persona description extracted using the Persona Extractor from PESConv \cite{cheng-etal-2023-pal}. 
To preserve topical coherence, entries are grouped by problem type, and retrieval is restricted within the same category.

\textbf{Query Construction.}
Given the current help-seeker utterance $u_t$ and persona description $p_t$, we construct the query as:
\begin{equation}
q = [p_t \; \text{[SEP]} \; u_t].
\end{equation}
Each candidate entry is represented as:
\begin{equation}
x_i = [p_i \; \text{[SEP]} \; s_i \; \text{[SEP]} \; r_i].
\end{equation}

\textbf{Dual-Encoder Retrieval.}
We adopt a Dense Passage Retrieval~\cite{rajapakse2023dense-dpr} model implemented with a dual-encoder architecture, consisting of a query encoder $f_q(\cdot)$ and a passage encoder $f_x(\cdot)$, both initialized from \textit{bert-base-uncased}~\cite{devlin-etal-2019-bert}.

For each candidate $x_i$, two similarity scores are computed:
\begin{equation}
\text{sim}_{ctx}(q, x_i)
= \text{sim}\big(f_q(q),\, f_x(x_i)\big),
\end{equation}
which captures contextual semantic relevance, and
\begin{equation}
\text{sim}_{per}(p_t, p_i)
= \text{sim}\big(f_q(p_t),\, f_x(p_i)\big),
\end{equation}
which measures persona-level consistency. The final retrieval score is defined as:
\begin{equation}
\text{sim}(i)
= \alpha \cdot \text{sim}_{ctx}(q, x_i)
+ \beta \cdot \text{sim}_{per}(p_t, p_i),
\end{equation}
where $\alpha$ and $\beta$ control the relative contributions of contextual and persona relevance.

\textbf{Demonstration Construction.}
We select the top-$k$ candidate pairs according to $\text{sim}(i)$ and construct structured demonstrations as:
\begin{align}
d_i &=
\text{[User: } u_i \text{]}
+ \text{[Persona: } p_i \text{]} \notag \\
&\quad
+ \text{[STRATEGY: } s_i \text{]}
+ \text{[SYSTEM: } r_i \text{]}, \label{eq:di} \\[0.3em]
D &=
d_1 \oplus d_2 \oplus \dots \oplus d_k, \label{eq:D} \\[0.3em]
H_P &=
\text{Encoder}(D), \label{eq:hp}
\end{align}
where $\oplus$ denotes separator-aware concatenation. The retrieved prompt $D$ is prepended to the dialogue context to guide response generation, with its length capped at 512 tokens for efficiency. By incorporating both semantic and persona-aware matching, the PR module retrieves demonstrations that are contextually relevant and stylistically consistent.

\subsection{Causality-aware Cognitive Filtering}
We propose a CCF module, which enhances emotional reasoning by filtering irrelevant commonsense knowledge, identifying causally salient context, and refining cognitive representations in a unified framework.

\textbf{Commonsense Filter.}
We first encode the dialogue context by concatenating all utterances with speaker-aware separators:
\begin{equation}
H_{\text{CTX}} = \text{Encoder}(u_1 \oplus u_2 \oplus \cdots \oplus u_N),
\end{equation}
where $\oplus$ denotes utterance-level concatenation.

We consider four cognitively meaningful relations:
\begin{equation}
r \in \mathcal{R} = \{\textit{xWant}, \textit{xNeed}, \textit{xIntent}, \textit{xEffect}\}.
\end{equation}

For each relation $r$, COMET \cite{bosselut-etal-2019-comet} generates $k$ commonsense inferences:
\begin{equation}
COM^r = \{ com_1^r, com_2^r, \dots, com_k^r \},
\end{equation}
where $k$ follows the default commonsense generation configuration.

To filter irrelevant inferences, we employ a DeBERTa-v3-large classifier trained on ComFact:
\begin{equation}
\hat{y}_i^r = \text{DeBERTa}_{Filter}(com_i^r),
\end{equation}
where $\hat{y}_i^r \in \{\texttt{RR},\texttt{IR}\}$. 
Only relevant candidates are retained:
\begin{equation}
COM_{\text{filter}}^r
= \mathop{\oplus}\limits_{i:\hat{y}_i^r=\texttt{RR}} com_i^r .
\end{equation}

The filtered sequence is encoded as:
\begin{equation}
E_{\text{com}}^{r} = \text{Encoder}(COM_{\text{filter}}^r),
\end{equation}
and all relations are aggregated:
\begin{equation}
E_{\text{com}}^{all} =
E_{\text{com}}^{\text{Intent}} \oplus
E_{\text{com}}^{\text{Want}} \oplus
E_{\text{com}}^{\text{Need}} \oplus
E_{\text{com}}^{\text{Effect}}.
\end{equation}

\textbf{Cause-aware Context Extractor.}
We adopt an external Emotion Cause Detector (ECD) \cite{poria2021causaldetector}, trained on RECCON, to identify utterances contributing to the emotion expressed in the current turn. 
Given the dialogue context, target utterance $u_t$, and emotion label $e_t$, the detector assigns:
\begin{equation}
y_i = f_{\text{cause}}(u_i, u_t, e_t), \quad y_i \in \{0,1\},
\end{equation}
where $y_i=1$ indicates an emotion-cause utterance.

Tokens from causal utterances are marked to construct a token-level causal attention mask:
\begin{equation}
{M}_{\text{causal}} \in \{0,1\}^{T \times T},
\end{equation}
which restricts attention to causally relevant spans. 
The original context representation is preserved in parallel and reintroduced during multi-source fusion.

\textbf{Cognitive Knowledge Refinement.}
Filtered commonsense knowledge is processed by a \textit{Cognitive Encoder} with the causal mask:
\begin{equation}
H_{\text{com}} = \text{Enc}_{\text{cog}}(E_{\text{com}}^{all},\, {M}_{\text{causal}}).
\end{equation}

The \texttt{[CLS]} hidden state serves as a cognitive summary:
\begin{equation}
h_{\text{com}} = H_{\text{com}}[0].
\end{equation}

This summary is repeated and concatenated with the context embedding:
\begin{equation}
H_{\text{mix}} = h_{\text{com}}^{\uparrow} \oplus H_{\text{CTX}},
\end{equation}
and refined using a causal-aware transformer:
\begin{equation}
H_{\text{ref}} = \text{Enc}_{\text{ref}}(H_{\text{mix}},\, {M}_{\text{causal}}).
\end{equation}

Finally, a \textit{Cognitive Selector} applies gated feature selection to obtain the final representation:
\begin{equation}
H_{C} = \text{MLP}(\sigma(H_{\text{ref}}) \odot H_{\text{ref}}),
\end{equation}
which serves as the cause-aware contextual representation for the decoder.

\subsection{Multi-Source Fusion for Generation}

To integrate heterogeneous information sources in PRCCF, we introduce a MSFG that combines three representations: the dialogue context encoding $H_{\text{CTX}}$, the retrieved demonstration representation $H_P$, and the cognitive representation $H_C$. This fusion provides the decoder with unified contextual, personalized, and causality-aware signals.

We compute context-aware demonstration and cognitive representations by attending to $H_P$ and $H_C$ using the dialogue context:
\begin{align}
Z_P &= \text{Softmax}(H_{\text{CTX}} \cdot H_P^\top) \cdot H_P, \\
Z_C &= \text{Softmax}(H_{\text{CTX}} \cdot H_C^\top) \cdot H_C.
\end{align}

The attended representations are fused with the original context via residual normalization:
\begin{align}
\tilde{H}_P^{\text{CTX}} &= \text{LayerNorm}(H_{\text{CTX}} + Z_P), \\
\tilde{H}_C^{\text{CTX}} &= \text{LayerNorm}(H_{\text{CTX}} + Z_C).
\end{align}

Similarly, knowledge-to-context alignment is performed to derive the complementary representations $\hat{H}_P^{\text{CTX}}$ and $\hat{H}_C^{\text{CTX}}$.

All representations are then combined through a learnable weighted fusion:
\begin{align}
H_{\text{fin}} &=
\lambda_1 \cdot H_{\text{CTX}}
+ \lambda_2 \cdot \tilde{H}_P^{\text{CTX}}
+ \lambda_3 \cdot \tilde{H}_C^{\text{CTX}} \notag \\
&\quad
+ \lambda_4 \cdot \hat{H}_P^{\text{CTX}}
+ \lambda_5 \cdot \hat{H}_C^{\text{CTX}}, \label{eq:hfin} \\[0.3em]
\lambda_i &=
\frac{e^{w_i}}{\sum_j e^{w_j}}, \label{eq:lambda}
\end{align}
where the weights $w_i$ are trainable and enable adaptive contribution from each source.

The fused representation is normalized before decoding:
\begin{equation}
\hat{H}_{\text{fin}} = \text{LayerNorm}(H_{\text{fin}}).
\end{equation}

Response generation is conditioned on the fused state. At each decoding step $t$, the next-token probability is computed as:
\begin{equation}
P(y_t \mid y_{<t}, C) = \text{Decoder}(E_{y_{<t}}, \hat{H}_{\text{fin}}),
\end{equation}
and the model is trained by minimizing the negative log-likelihood:
\begin{equation}
\mathcal{L} = -\frac{1}{n} \sum_{t=1}^{n} \log P(y_t \mid C, y_{<t}).
\end{equation}

\begin{table*}[htbp]
  \centering
  \caption{Results of automatic evaluation on ESConv, where the values of baselines are sourced from public papers. The bolded and underlined results indicate best and second-best result respectively.}
  \normalsize
  \resizebox{\textwidth}{!}{
  \begin{tabular}{lccccccccc}
    \hline
    \textbf{Model} & \textbf{ACC(\%)$\uparrow$} & \textbf{PPL$\downarrow$} & \textbf{B-1$\uparrow$}& \textbf{B-2$\uparrow$}& \textbf{B-3$\uparrow$} & \textbf{B-4$\uparrow$} & \textbf{D-1$\uparrow$} & \textbf{D-2$\uparrow$} & \textbf{R-L$\uparrow$}\\
    \hline
    \textbf{MIME}(2020 EMNLP)\cite{majumder-etal-2020-mime} &- &43.27 &16.15 &4.82 &1.79 &1.03 &2.56 &12.33 &14.83  \\
    \textbf{BlenderBot-Joint}(2021 ACL)\cite{liu-2021-esconv} &27.72 &18.11 &16.99 &6.18 &2.95 &1.66 &3.27 &20.87 &15.13  \\
    \textbf{MISC}(2022 ACL)\cite{tu-etal-2022-misc} &31.63 &16.16 &- &7.31 &- &2.20 &4.41 &19.71 &17.91  \\
    \textbf{GLHG}(2022 IJCAI)\cite{peng2022glhg} &- &15.67 &19.66 &7.57 &3.74 &2.13 &3.50 &21.26 &16.37 \\
    \textbf{KEMI}(2023 ACL)\cite{deng-etal-2023-knowledge-KEMI} &- &15.92 &- &8.31 &- &2.51 &- &- &17.05 \\
    \textbf{FADO}(2023 Knowledge-Based Systems)\cite{Peng23-FADO} &32.41 &15.52 &- &8.31 &4.36 &2.66 &3.80 &21.39 &18.09  \\
    \textbf{SUPPORTER}(2023 ACL)\cite{zhou-etal-2023-supporter} &- &15.37 &19.50 &7.49 &3.58 &- &4.93 &27.73 &-  \\
    \textbf{TransESC}(2023 ACL)\cite{zhao-etal-2023-transesc} &34.71 &15.85 &17.92 &7.64 &4.01 &2.43 &4.73 &20.48 &17.51 \\
    \textbf{PAL}(2023 ACL)\cite{cheng-etal-2023-pal} &34.51 &15.92 &- &8.75 &- &2.66 &5.00 &\underline{30.27} &18.06  \\
    \textbf{CauESC}(2024)\cite{chen2024CauESC} &33.33 &15.30 &- &8.17 &4.55 &2.82 &4.70 &19.85 &18.20 \\
    \textbf{MFF-ESC}(2024 IREC)\cite{bao-etal-2024-MFFESC} &- &16.43 &20.64 &8.87 &4.81 
    &\underline{2.98} &\underline{5.34} &22.18 &18.83 \\
    \textbf{DQ-HGAN}(2024 Knowledge-Based Systems)\cite{LI2024-DQ-HGAN} &- &15.78 &- &8.53 &3.79 &2.25 &5.25 &21.98 &\underline{19.56} \\
    \textbf{D$^2$RCU}(2024 SIGIR)\cite{xu24-ddrcu} &35.32 &15.43 &21.20 &9.01 &4.81 &2.94 &4.97 &26.21 &18.71 \\
    \textbf{DKPE}(2025 coling)\cite{hao-kong-2025-DKPE} &\underline{35.51} &\underline{14.88} &\underline{21.38} &\underline{9.27} &\underline{4.93} &2.92 &4.88 &25.95 &18.87 \\
    \hline
    \textbf{PRCCF}(ours) &\textbf{40.72} &\textbf{13.10} &\textbf{23.94} &\textbf{10.71} &\textbf{5.87} &\textbf{3.55} &\textbf{6.17} &\textbf{31.16} &\textbf{19.78} \\
    \hline
  \end{tabular}
}
  \label{tab:baselineresult}
\end{table*}
\begin{table}[htbp]
  \centering
  \caption{Comparisons with Large Language Models, where the values of large language model baselines are sourced from public papers.}
  \resizebox{\columnwidth}{!}{\begin{tabular}{lcc}
    \hline
    \textbf{Model} &\textbf{B-2$\uparrow$} &\textbf{R-L$\uparrow$}\\
    \hline
    \textbf{ChatGPT 1-shot}\cite{zhao2023chatgptoneshot} &4.53 &13.19 \\
    \textbf{ChatGLM-6B w/P-Tuning}\cite{bao-etal-2024-MFFESC} &7.22 &\underline{16.15} \\
    \textbf{ChatGPT w/Example Expansion}\cite{kang-etal-2024-chatgptee} &\underline{7.45} &15.22\\
    \textbf{FSM}\cite{zhao2025FSM} &3.25 &15.80\\
    \textbf{straQ$^*$-distill}\cite{wang2025straQ} &6.25 &15.44\\
    \hline
    \textbf{PRCCF} &\textbf{10.71} &\textbf{19.78} \\
    \hline
  \end{tabular}
}
  \label{tab:llm}
\end{table}
\section{Experiments}
\subsection{Datasets}
We evaluate our approach on the Emotional Support Conversation dataset (ESConv), a high-quality crowdsourced benchmark widely used in ESC research \cite{cheng-etal-2023-pal,zhao-etal-2023-transesc,Peng23-FADO,xu24-ddrcu,chen2024CauESC,hao-kong-2025-DKPE}. 
ESConv contains 1,300 multi-turn conversations with an average of 29.8 turns per dialogue, making it substantially more interactive than standard empathic dialogue datasets \cite{rashkin-2019-empatheticdataset}. 
In each turn, supporters generate responses (average length: 20.2 tokens) using one of eight predefined support strategies. 
Following \citet{liu-2021-esconv}, the dataset is split into training (80\%), validation (10\%), and test (10\%) sets.

\subsection{Evaluation Metrics}
\textbf{Automatic Evaluation} 
We adopt a comprehensive set of automatic metrics, including strategy prediction accuracy (ACC), perplexity (PPL), BLEU-n (B-n) \cite{papineni-etal-2002-bleu}, Distinct-n (D-n) \cite{li-etal-2016-Dist-n}, and ROUGE-L (R-L) \cite{lin-2004-rouge-R-L}. 
ACC measures the correctness of predicted support strategies, while PPL reflects the model’s confidence in generation. 
B-n and R-L evaluate the overlap between generated and reference responses, and D-n assesses response diversity by the proportion of unique n-grams.

\textbf{Human A/B Evaluation} 
Following \citet{liu-2021-esconv}, we conduct human evaluations on 100 randomly sampled test dialogues. 
Crowd workers interact with two models under the same scenario and assess the responses along four dimensions: Identification, Comforting, Suggestion, and Overall Preference.

\subsection{Training Details}
For fair comparison, we adopt the same backbone as \citet{liu-2021-esconv}, based on BlenderBot-small \cite{roller-etal-2021-blednerbot}. The maximum encoder and decoder input lengths are set to 512 and 50, respectively. Retrieved demonstrations and cognitive sequences are both truncated to 512 tokens.

The model is trained on a single NVIDIA 3090Ti GPU using Adam optimizer ($\beta_1=0.9$, $\beta_2=0.999$) with an initial learning rate of $1.5 \times 10^{-5}$. The training and validation batch sizes are 8 and 16. Training runs for up to 10 epochs, and the checkpoint with the lowest validation perplexity is selected. During inference, we employ a sampling-based decoding strategy with top-$k=10$, top-$p=0.9$, and a repetition penalty of $1.03$.

\subsection{Baselines}

We compare PRCCF with a wide range of competitive ESC baselines, including:
MIME \cite{majumder-etal-2020-mime},
BlenderBot-Joint \cite{liu-2021-esconv},
GLHG \cite{peng2022glhg},
MISC \cite{tu-etal-2022-misc},
KEMI \cite{deng-etal-2023-knowledge-KEMI},
FADO \cite{Peng23-FADO},
SUPPORTER \cite{zhou-etal-2023-supporter},
TransESC \cite{zhao-etal-2023-transesc},
PAL \cite{cheng-etal-2023-pal},
CauESC \cite{chen2024CauESC},
D$^2$RCU \cite{xu24-ddrcu},
MFF-ESC \cite{bao-etal-2024-MFFESC},
DKPE \cite{hao-kong-2025-DKPE}.

We also include several LLM-based baselines:
ChatGPT (1-shot) \cite{zhao2023chatgptoneshot},
ChatGLM-6B with P-Tuning \cite{du-etal-2022-chatglm,liu-2021-esconv},
ChatGPT with Example Expansion \cite{kang-etal-2024-chatgptee},
FSM \cite{zhao2025FSM},
straQ$^*$-distill \cite{wang2025straQ}.

\begin{table}[htbp]
  \centering
  \caption{The results of ablation study for PRCCF.}
  \resizebox{\columnwidth}{!}{\begin{tabular}{lcccccc}
    \hline
    \textbf{Model} &\textbf{PPL$\downarrow$} &\textbf{B-1$\uparrow$} &\textbf{B-2$\uparrow$}&\textbf{B-3$\uparrow$} &\textbf{B-4$\uparrow$} &\textbf{R-L$\uparrow$}\\
    \hline
    \textbf{PRCCF} &\textbf{13.10} &\textbf{23.94} &\textbf{10.71} &\textbf{5.87} &\textbf{3.55} &\textbf{19.78} \\
    w/o PR &13.19 &17.27 &6.95 &3.53 &2.05 &16.76 \\
    w/o Per$_{sim}$ &13.28 &22.89 &9.96 &5.37 &3.22 &19.13 \\
    w/o CCF &14.76 &20.49 &8.72 &4.59 &2.71 &17.63 \\
    w/o Causal &13.27 &22.16 &9.59 &5.13 &3.05 &19.01 \\
    w/o Filter &14.59 &21.70 &9.60 &5.18 &3.10 &18.07 \\
    \hline
  \end{tabular}
}
  \label{tab:ablation}
\end{table}

\begin{figure}[ht]
  \centering
  \includegraphics[width=\linewidth]{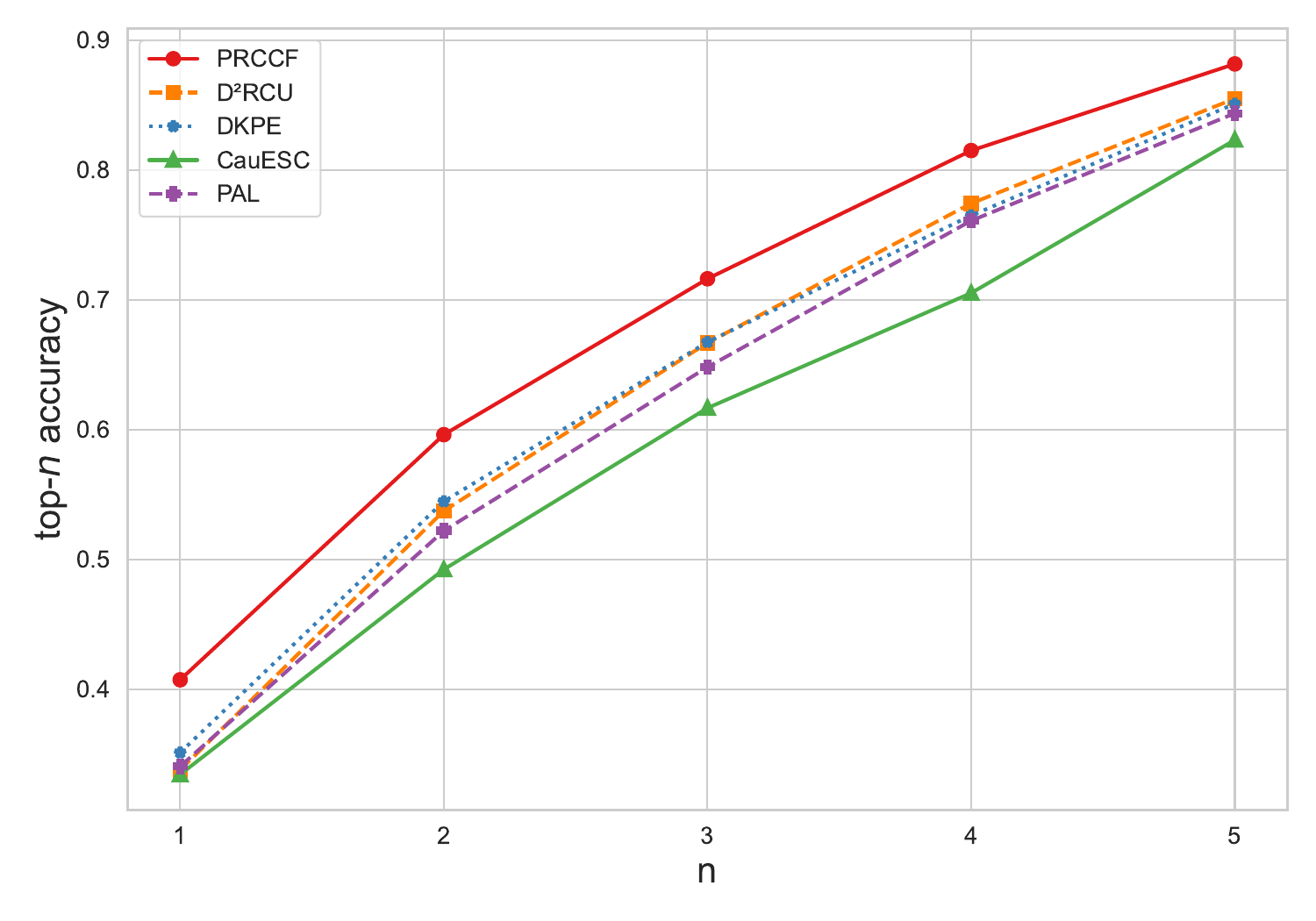} 
  \caption{Top-$n$ accuracy comparison across different methods.}
  \label{fig:topnacc}
\end{figure}

\section{RESULTS AND ANALYSIS}
\subsection{Automatic Evaluation}

Table \ref{tab:baselineresult} reports the automatic evaluation results on the ESConv dataset, where baseline scores are taken from their original publications for fair comparison. PRCCF achieves the best overall performance across most metrics, demonstrating its effectiveness in both response generation and strategy modeling. Specifically, PRCCF obtains the lowest perplexity, indicating more stable and coherent language generation, and achieves the highest strategy prediction accuracy, reflecting a stronger alignment between generated responses and underlying supportive intents. In addition, PRCCF consistently outperforms existing baselines on BLEU scores and shows competitive performance on diversity and semantic overlap metrics, suggesting that it can generate fluent, diverse, and contextually appropriate emotional support responses.

Table \ref{tab:llm} presents the comparison between PRCCF and representative large language model based baselines, with all results taken from their original publications. While LLM-based approaches such as ChatGPT with example expansion and ChatGLM-6B with P-Tuning show strong performance on individual metrics, PRCCF substantially surpasses all LLM baselines on both BLEU-2 and ROUGE-L. These results indicate that general-purpose large language models, despite their strong generative capabilities, are less effective at capturing fine-grained emotional needs and structured support strategies in ESC. By explicitly incorporating persona-guided retrieval and causality-aware cognitive filtering, PRCCF is better able to leverage personalized demonstrations and cognitively aligned knowledge, leading to more accurate lexical grounding and more coherent emotional support responses.
\begin{figure*}[htbp]
    \centering
    \includegraphics[width=0.9\textwidth]{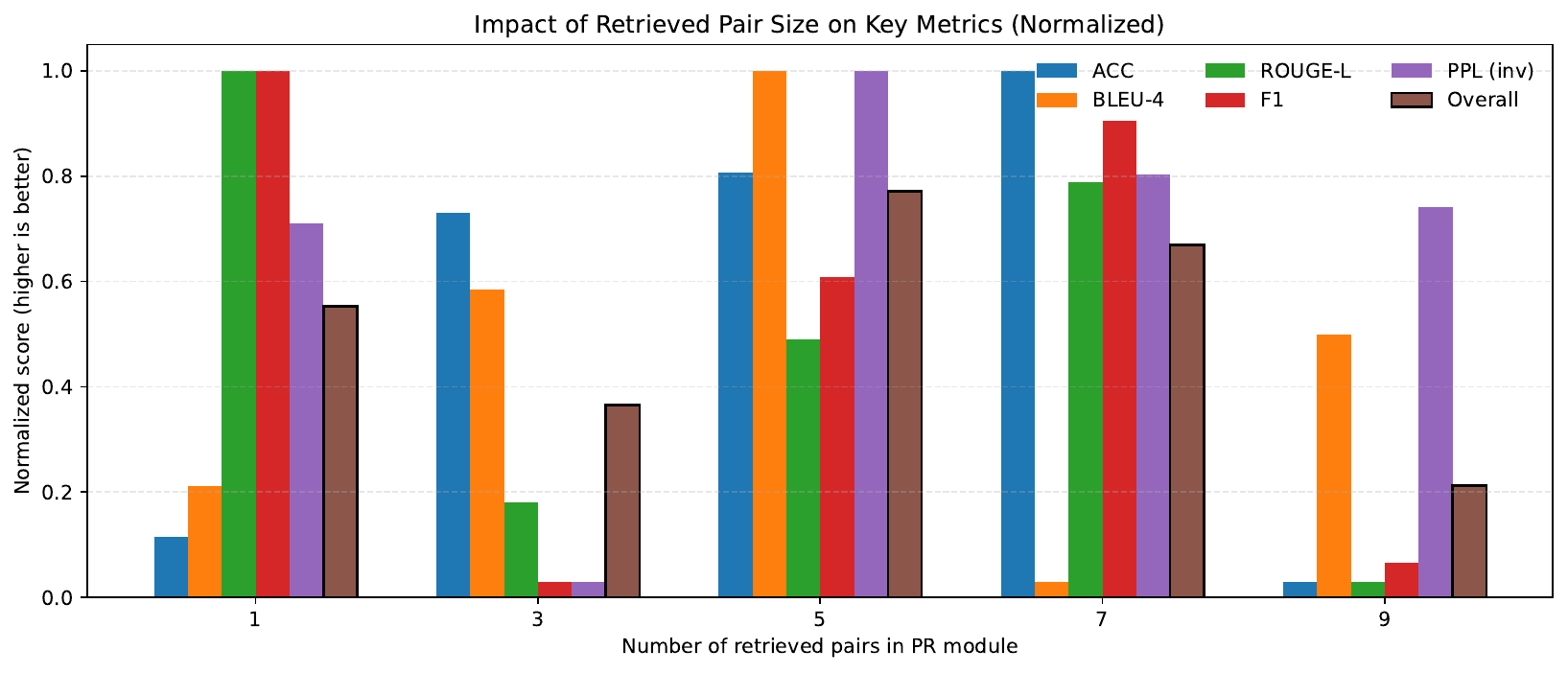}
    \caption{Effect of the number of retrieved candidate pairs (\textit{pairs}) in the PR module. All metrics are min--max normalized, with perplexity inverted so that higher values indicate better fluency. Performance peaks at $\textit{pairs}=5$.}
    \label{fig:pairs_ablation}
\end{figure*}

\begin{table}[htbp]
\centering
\caption{The results of the human evaluation (\%). 
$\dagger$ and $\ddagger$ indicate improvements with p-value $<0.1$ and p-value$<0.05$, respectively. 
Significance is tested using a one-sided exact sign test on non-tied cases (ties are excluded).}
\resizebox{\columnwidth}{!}{
\begin{tabular}{llccc}
\hline
\textbf{Comparisons} & \textbf{Aspects} & \textbf{Win} & \textbf{Lose} & \textbf{Tie} \\
\hline
\multirow{4}{*}{Ours vs. D$^2$RCU}
& Identification & \textbf{58}$^{\ddagger}$ & 32 & 10 \\
& Comforting     & \textbf{60}$^{\ddagger}$ & 33 & 7  \\
& Suggestion     & \textbf{47} & 36 & 17 \\
& Overall        & \textbf{57}$^{\ddagger}$ & 37 & 6  \\
\hline
\multirow{4}{*}{Ours vs. DKPE}
& Identification & \textbf{47} & 36 & 17 \\
& Comforting     & \textbf{50}$^{\dagger}$ & 37 & 13 \\
& Suggestion     & \textbf{46}$^{\ddagger}$ & 28 & 26 \\
& Overall        & \textbf{51}$^{\dagger}$ & 37 & 12 \\
\hline
\end{tabular}
}
\label{tab:humaneval}
\end{table}

\subsection{Ablation Study}

We conduct ablation studies to evaluate the contributions of the PR and CCF modules, with results reported in Table \ref{tab:ablation}. Removing the PR module (w/o PR) causes the most severe performance drop, particularly on BLEU and R-L, demonstrating that retrieval-based demonstrations are essential for maintaining response coherence and emotional consistency. Removing persona-aware similarity (w/o Per$_{sim}$) also leads to consistent declines across BLEU metrics, confirming the importance of persona alignment in demonstration retrieval.
\begin{figure}[htbp]
    \centering
    \includegraphics[width=\linewidth]{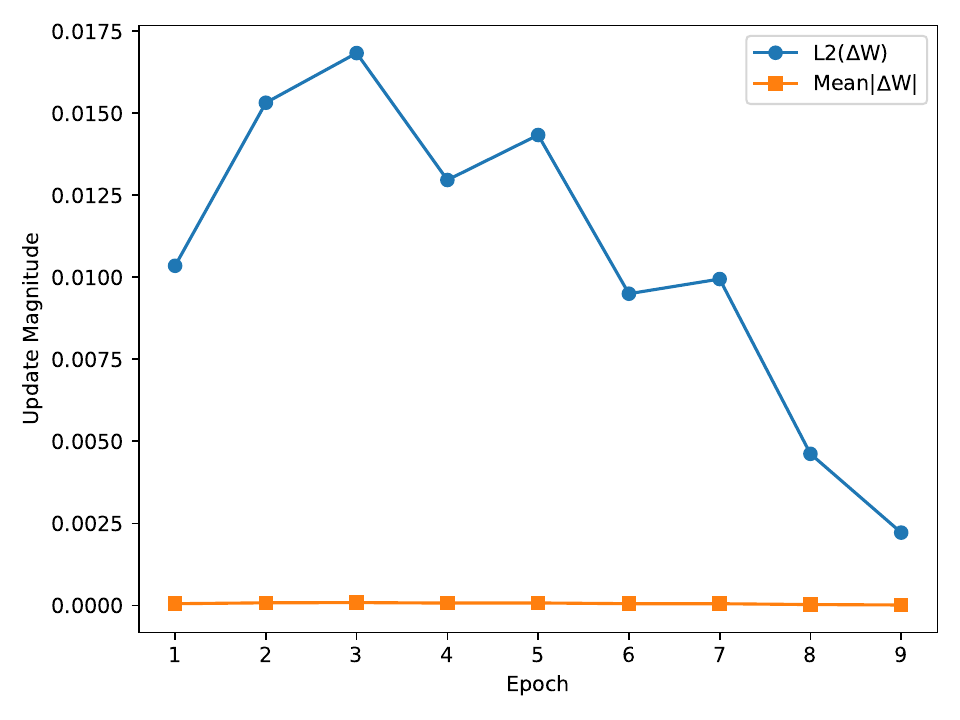}
    \caption{Parameter update dynamics of the Cognitive Knowledge Refinement across training epochs.}
    \label{fig:cog_refinement_update}
\end{figure}

For the CCF module, removing it entirely (w/o CCF) results in the highest perplexity, indicating that unfiltered commonsense knowledge introduces noise and harms fluency. Disabling cause-aware context modeling (w/o Causal) mainly affects higher-order BLEU scores, while removing commonsense filtering (w/o Filter) leads to the largest degradation among CCF variants, highlighting the necessity of filtering irrelevant knowledge. Overall, these results show that PR and CCF contribute complementary benefits to emotionally grounded and coherent response generation.

\begin{table*}[htbp]
  \centering
  \caption{Case study of the generated responses by baselines and PRCCF.}
  \label{tab:casestudy1}
  \resizebox{\textwidth}{!}{
    \begin{tabular}{ll}
      \hline 
      \multicolumn{2}{c}{\textbf{Situation}} \\  
      \hline  
      \multicolumn{2}{c}{My job just laid off 10\% of the workforce and I'm afraid I'm next.} \\
      \hline
      \multicolumn{2}{c}{\textbf{Context}} \\
      \hline
      Seeker & My company just laid off 10\% of the workforce and furloughed another 10\%. I'm worried I'm going to be in the next round. \\
      Supporter & Why do you think you are the next round? Do you know how did they choose them? \\
      Seeker & Just based on their tenure and job experience. I'm around the same level. \\
      Supporter & I know, this year was stressful for everyone in our planet, so you are not alone... \\
      Help-Seeker & \textbf{If I'm laid off, we can't afford our home. We are already dipping into savings because my wife lost her job.} \\
      \hline
      \multicolumn{2}{c}{\textbf{Response}} \\
      \hline
      Ground-Truth & Did you try searching for new jobs? \\
      D$^2$RCU & I understand, it's a tough situation for everyone. I would suggest moving to another location, maybe in your city? \textit{(Irrelevant advice ignoring core anxiety)} \\
      DKPE & That is a very hard situation. I know, it is hard to know what will happen if you are not the next round of hiring. \textit{(Confusing cognition and out of context)} \\
      \textbf{PRCCF} & \textbf{I understand, this year has been a very bad year for everyone, do you want to see other choices? \textit{(Empathy and effective support)}} \\
      \hline  
    \end{tabular}
  }
\end{table*}

\subsection{Human Evaluation}

Table \ref{tab:humaneval} reports the results of the human evaluation. Compared with D$^2$RCU, PRCCF achieves more Win cases across all metrics, with 58, 60, and 47 wins on Identification, Comforting, and Suggestion, respectively, and 57 wins on the Overall metric. This indicates a stronger ability to understand users’ emotional states and provide appropriate emotional support and suggestions.

PRCCF also consistently outperforms DKPE, achieving 47, 50, and 46 wins on the three fine-grained metrics and 51 wins on the Overall metric. These results show that PRCCF produces emotionally supportive and intention-aligned responses with more helpful suggestions.

\subsection{Case Study}
Table~\ref{tab:casestudy1} presents a representative ESConv test case involving strong anxiety about potential job loss. We compare responses generated by D$^2$RCU, DKPE, and PRCCF to qualitatively assess emotional alignment and contextual appropriateness.

D$^2$RCU provides a generic response that overlooks the seeker’s financial distress and offers advice with limited contextual relevance. DKPE produces a more fluent reply, but its content remains emotionally misaligned and risks exacerbating anxiety by introducing confusing or inappropriate guidance.

In contrast, PRCCF accurately identifies the seeker’s core emotional concerns and responds with validation and reassurance rather than premature problem-solving. By prioritizing emotional stabilization and situational awareness, PRCCF delivers support that is more consistent with counseling principles. This case demonstrates how persona-guided retrieval and causality-aware filtering contribute to context-sensitive and emotionally aligned responses.
\begin{figure*}[htbp]
  \centering
  \includegraphics[width=0.95\textwidth]{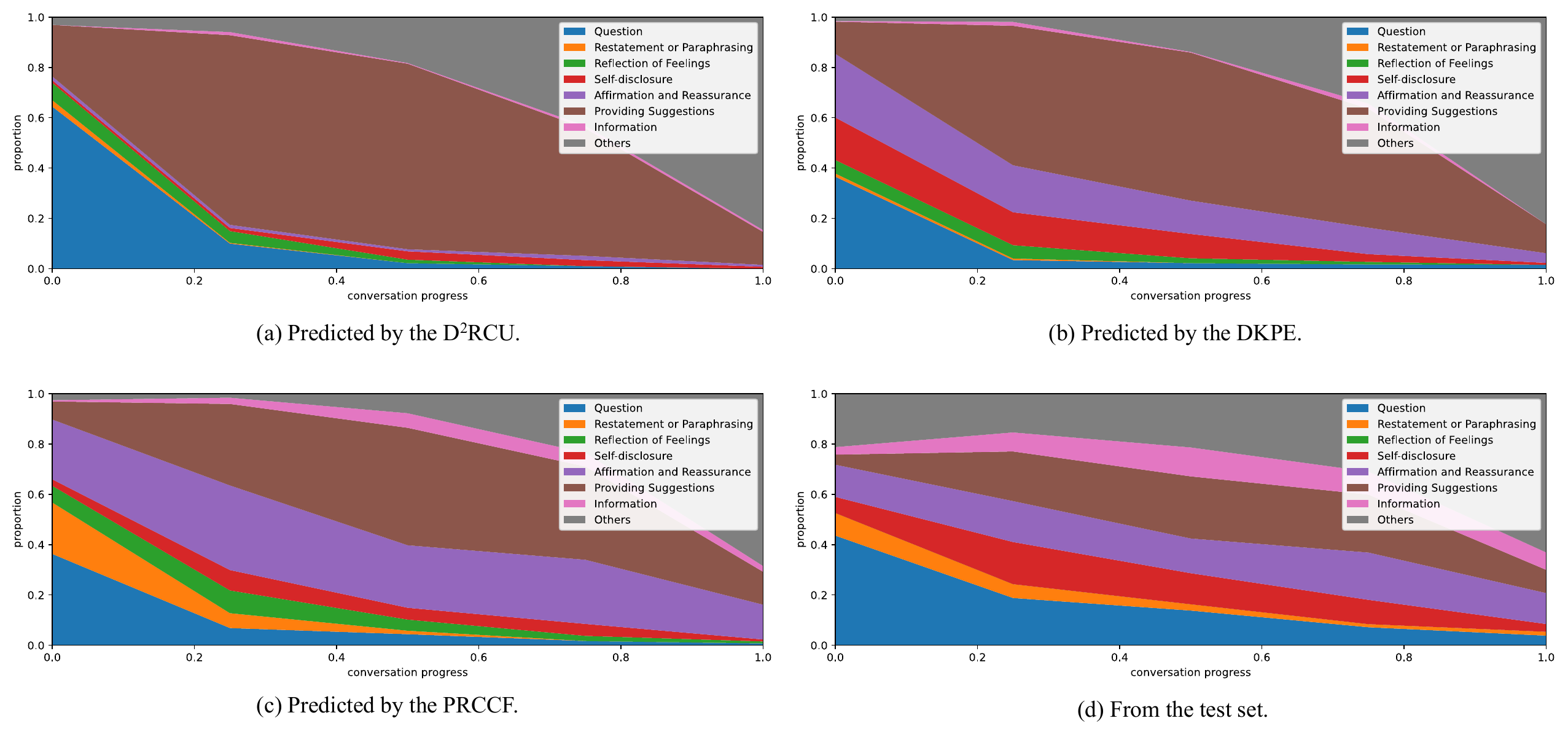} 
  \caption{The strategy distribution in the different stage of conversation.}
  \label{fig:strat}
\end{figure*}
\subsection{Top-$n$ Strategy Prediction Accuracy}
To investigate whether PRCCF enhances the model’s ability to anticipate appropriate support strategies, we report the top-$n$ strategy prediction accuracy across competing systems. For each dialogue turn, a model is required to rank all candidate strategies, and a prediction is counted as correct if the gold label appears within the top-$n$ positions.

As shown in Figure~\ref{fig:topnacc}, PRCCF consistently outperforms D$^2$RCU, DKPE, CauESC, and PAL across all values of $n$. The performance margins are especially notable for $n=1$ and $n=2$, indicating that PRCCF more accurately captures both the seeker’s evolving emotional needs and the contextual cues that govern strategy selection. Accuracy continues to increase as $n$ grows, but PRCCF maintains the leading position throughout, demonstrating that both its retrieval mechanism and its causality-aware cognitive filtering contribute to more reliable strategy reasoning. These results highlight the advantage of incorporating personalized demonstrations and refined cognitive cues when modeling strategy decisions in emotional support dialogue.

\subsection{Analysis of Cognitive Knowledge Refinement}
To verify that the proposed Cognitive Knowledge Refinement module is effectively optimized during training, we analyze the parameter update dynamics of the refinement process that integrates cognitive knowledge with contextual representations under causal constraints. As shown in Figure~\ref{fig:cog_refinement_update}, the component exhibits non-zero parameter updates throughout training, with larger updates in early epochs followed by a gradual decay as training converges. This pattern indicates that the refinement mechanism consistently receives optimization signals and undergoes stable adaptation rather than remaining static. Although the mean absolute update is small in magnitude, it remains consistently above zero, reflecting fine-grained and controlled parameter refinement. Overall, these observations confirm that the Cognitive Knowledge Refinement is properly optimized and actively participates in the training process.

\subsection{Effect of Retrieved Pair Size}

We investigate the influence of the number of retrieved candidate pairs in the PR module by varying the value of $\textit{pairs}$ and reporting both strategy classification accuracy and generation quality. As illustrated in Figure~\ref{fig:pairs_ablation}, performance improves steadily when increasing $\textit{pairs}$ from 1 to 5. In particular, B-4, R-L, and F1 exhibit consistent gains, and the aggregated overall score reaches its maximum at $\textit{pairs}=5$, suggesting more comprehensive semantic coverage and better demonstration guidance.

However, when the number of retrieved pairs increases beyond 5, the benefits diminish. Settings with $\textit{pairs}=7$ and $\textit{pairs}=9$ yield no further performance gains, and several metrics decline. This indicates that retrieving too many pairs introduces less relevant or noisy demonstrations, which weakens the model’s ability to focus on the most informative examples.

Based on these observations, we adopt $\textit{pairs}=5$ as the default configuration for all subsequent experiments, as it provides a balanced trade-off between semantic diversity and noise control.

\subsection{Strategy Distribution Analysis}
Following the three-stage supportive process~\cite{hill2013helping}, consisting of \textit{Exploration}, \textit{Comforting}, and \textit{Action}, we analyze strategy evolution on the test set by dividing each dialogue into six equal intervals.
Figure~\ref{fig:strat} compares the ground-truth and predicted strategy distributions across these intervals.

PRCCF exhibits a clear three-stage progression. In early intervals, it mainly adopts exploratory strategies such as Question and Reflection of Feelings. As the dialogue progresses, Affirmation and Reassurance become more frequent, indicating a shift toward emotional validation. In later intervals, Providing Suggestions dominates, reflecting a transition to action-oriented support.

Compared with D$^2$RCU and DKPE, PRCCF more consistently captures this gradual strategy shift across dialogue stages, aligning more closely with Hill’s theoretical process. This suggests that PRCCF better models the temporal dynamics of emotional support conversations.

\section{Conclusion}
We propose PRCCF, a Persona-guided Retrieval and Causality-aware Cognitive Filtering framework for Emotional Support Conversations. By jointly modeling persona-consistent demonstration retrieval and causally grounded cognitive knowledge, PRCCF effectively captures both personalized user needs and underlying emotional causes, enabling more context-sensitive and psychologically informed responses.

Experiments on the ESConv dataset show that PRCCF consistently outperforms state-of-the-art baselines on automatic and human evaluations, demonstrating improved empathy, coherence, and support quality. 

Future work will explore richer user modeling and finer-grained cognitive representations, as well as integrating PRCCF with reinforcement learning and large language models to enhance adaptability and scalability.
\bibliography{tacl2021}
\bibliographystyle{acl_natbib}

\iftaclpubformat

\onecolumn

% \appendix
% \section{Author/Affiliation Options as set forth by MIT Press}
% \label{sec:authorformatting}

% Option 1. Author’s address is underneath each name, centered.

% \begin{quote}\centering
%   \begin{tabular}{c}
%     \textbf{First Author} \\
%     First Affiliation \\
%     First Address 1 \\
%     First Address 2 \\
%     \texttt{first.email@example.com}
%   \end{tabular}
%   \ 
%   \begin{tabular}{c}
%     \textbf{Second Author} \\
%     Second Affiliation \\
%     Second Address 1 \\
%     Second Address 2 \\
%     \texttt{second.email@example.com}
%   \end{tabular}

%   \begin{tabular}{c}
%     \textbf{Third Author} \\
%     Third Affiliation \\
%     Third Address 1 \\
%     Third Address 2 \\
%     \texttt{third.email@example.com}
%   \end{tabular}
% \end{quote}

% Option 2. Author’s address is linked with superscript characters to its name,
% author names are grouped, centered.

% \begin{quote}\centering
%     \textbf{First Author$^\diamond$} \quad \textbf{Second Author$^\dagger$} \quad
%     \textbf{Third Author$^\ddagger$}
%     \\ \ \\
%     $^\diamond$First Affiliation \\
%     First Address 1 \\
%     First Address 2 \\
%     \texttt{first.email@example.com}
%      \\ \ \\
%      $^\dagger$Second Affiliation \\
%     Second Address 1 \\
%     Second Address 2 \\
%     \texttt{second.email@example.com}
%      \\ \ \\
%     $^\ddagger$Third Affiliation \\
%     Third Address 1 \\
%     Third Address 2 \\
%     \texttt{third.email@example.com}
% \end{quote}
  
\fi

\end{document}